\documentclass[10pt,twocolumn,letterpaper]{article}
\usepackage{wacv}
\usepackage{times}
\usepackage{epsfig}
\usepackage{graphicx}
\usepackage{amsmath}
\usepackage{amssymb}
\usepackage{commath}
\usepackage{amsthm}
\usepackage{algorithmic}
\usepackage[super]{nth}
\usepackage[norelsize,longend,ruled,lined,boxed,commentsnumbered]{algorithm2e}
\usepackage{breqn}
\usepackage{bm}
\usepackage[accsupp]{axessibility}
\newtheorem{theorem}{Theorem}
\newtheorem{corollary}{Corollary}
\newtheorem{lemma}{Lemma}
 
\def\wacvPaperID{811} % Enter the WACV Paper ID here
\wacvalgorithmstrack   % Uncomment this line if you are submitting to the Algorithms Track.
\wacvfinalcopy % *** Uncomment this line for the final submission
\ifwacvfinal
\def\assignedStartPage{1}
\fi

\ifwacvfinal
\usepackage[breaklinks=true,bookmarks=false]{hyperref}
\else
\usepackage[pagebackref=true,breaklinks=true,colorlinks,bookmarks=false]{hyperref}
\fi

\ifwacvfinal
\else
\fi

\usepackage{fancyhdr}

\fancypagestyle{firstpage}{%
  \rhead{IEEE WACV 2023.}
  \lfoot{\small{This paper is accepted by IEEE/CVF Winter Conference on Applications of Computer Vision (WACV), 2023. Copyright © 2023 IEEE. Personal use of this material is permitted. However, permission to use this material for any other purposes must be obtained from the IEEE by sending an email to pubs-permissions@ieee.org.}}
}

\begin{document}
%%%%%%%%% TITLE
% \title{AdaNorm: Exponential Moving Gradient Norm Average based Adaptive Momentum for Convolutional Neural Networks}
% \title{AdaNorm: Gradient Norm Training History based Optimizer for Convolutional Neural Networks}
% \title{AdaNorm: Gradient Norm Training History based Optimizer for CNNs}
% \title{AdaNorm Optimizer using Training History based Gradient Norm Correction for Convolutional Neural Networks}
\title{AdaNorm: Adaptive Gradient Norm Correction based Optimizer for CNNs}

\author{Shiv Ram Dubey$^{\ast}$, Satish Kumar Singh$^{\ast}$, Bidyut Baran Chaudhuri$^{\dagger}$\\
$^{\ast}$Computer Vision and Biometrics Lab, Indian Institute of Information Technology, Allahabad\\
$^{\dagger}$Techno India University, Kolkata, India and Indian Statistical Institute, Kolkata, India\\
{\tt\small srdubey@iiita.ac.in, sk.singh@iiita.ac.in, bidyutbaranchaudhuri@gmail.com}
% \and
% Satish Kumar Singh\\
% Computer Vision and Biometrics Lab\\
% Indian Institute of Information Technology, Allahabad\\
% {\tt\small sk.singh@iiita.ac.in}
}

\maketitle
\thispagestyle{firstpage}

%%%%%%%%% ABSTRACT
\begin{abstract}
The stochastic gradient descent (SGD) optimizers are generally used to train the convolutional neural networks (CNNs). In recent years, several adaptive momentum based SGD optimizers have been introduced, such as Adam, diffGrad, Radam and AdaBelief. However, the existing SGD optimizers do not exploit the gradient norm of past iterations and lead to poor convergence and performance. In this paper, we propose a novel AdaNorm based SGD optimizers by correcting the norm of gradient in each iteration based on the adaptive training history of gradient norm. By doing so, the proposed optimizers are able to maintain high and representive gradient throughout the training and solves the low and atypical gradient problems. The proposed concept is generic and can be used with any existing SGD optimizer. We show the efficacy of the proposed AdaNorm with four state-of-the-art optimizers, including Adam, diffGrad, Radam and AdaBelief. We depict the performance improvement due to the proposed optimizers using three CNN models, including VGG16, ResNet18 and ResNet50, on three benchmark object recognition datasets, including CIFAR10, CIFAR100 and TinyImageNet. Code: \url{https://github.com/shivram1987/AdaNorm}.
\end{abstract}

%%%%%%%%% BODY TEXT
\section{Introduction}
In recent years, Convolutional Neural Networks (CNNs) have become the major parametric model to solve the Computer Vision problems \cite{gu2018recent}, such as Object Recognition \cite{vgg}, \cite{resnet}, Object Localization \cite{fasterrcnn}, \cite{yolo}, Image Segmentation \cite{maskrcnn}, Face Recognition \cite{facenet}, \cite{arcface}, Image Retrieval \cite{dubey2021decade}, Biomedical Image Analysis \cite{sirinukunwattana2016locality}, and many more. The training of CNN models is performed to learn the parameters of the network on the training set of data. 

In practice, the batch-wise Stochastic Gradient Descent (SGD) based optimization techniques are used to train the CNN models. The parameters/weights are first initialized using some approach, such as random initialization, Xavier initialization \cite{glorot2010understanding}, He initialization \cite{he2015delving}, etc. Then, the parameters are updated by utilizing the gradient of objective function w.r.t. the correspnding parameter in multiple iterations \cite{ruder2016overview}. The vanilla SGD approach \cite{sgd} updates the parameters in the opposite direction of gradient by a small step-size, called as learning rate. However, it suffers with various challenges, such as zero gradient at local minimum and saddle regions, severe variations in gradient in different directions, same step-size used for each parameter update irrespective of its behaviour, and bad gradient due to batch-wise computation. The SGD with Momentum (i.e., SGDM) \cite{sgdm} tackles the first two issues by considering the exponential moving average (EMA) of gradient for parameter update. The EMA of gradient builds the velocity in the direction of consistent gradient for faster convergence. The step-size problem is addressed by AdaGrad \cite{adagrad} which divides the step-size by the root of sum of past squared gradient. However, it leads to dying learning rate problem in the later stage of training, which is fixed in RMSProp \cite{rmsprop} by dividing the step-size with root of the EMA of squared gradient.
The Adam optimizer \cite{adam} combines the concept of SGDM and RMSProp and proposes adaptive moments. The first and second moments are computed as EMA of gradients and squared gradients, respectively. Adam uses first moment to update the parameters and second moment to control the step-size. Adam optimizer has been used successfully with various CNN models for different computer vision problems. In order to deal with the effect of bad batch-wise gradient on the effective learning rate the AMSGrad \cite{amsgrad} uses maximum of past squared gradients to control the learning rate, rather than exponential average. However, the AMSGrad does not deal with bad gradient used for parameter updates, which is taken care in the proposed AdaNorm optimizers.

The Adam optimizer suffers near the minimum due to high moment leading to overshooting of minimum and oscillation near minimum \cite{diffgrad}, \cite{radam}, \cite{adabelief}. Recent optimizers have tried to tackle this issue, such as diffGrad \cite{diffgrad} introduces a friction coefficient based on the local gradient behaviour to reduce the learning rate near minimum; Rectified Adam (i.e., Radam) \cite{radam} rectifies the variance of the adaptive learning rate and converts Adam into SGDM based on the variance threshold; and AdaBelief \cite{adabelief} considers the EMA of square of difference between the gradient and first order moment (i.e., belief information) to control the learning rate. The other variants of Adam includes Nostalgic Adam (NosAdam) \cite{nosadam} which gives more weight to the past gradients to incorporate the long-term memory. However, NosAdam miss to rectify the norm of the gradients. The AdaBound \cite{adabound} performs clipping to make the optimizer more robust to extreme learning rates, caused by adaptive momentum. The AdaBound approach can be seen as the post-correction of learning rates. Similarly, the adaptive and momental upper bounds are used in AdaMod \cite{adamod} to avoid the large learning rates in the initial iterations of Adam. The AdamP \cite{adamp} has shown that the decay in learning rate might lead to sub-optimal solution and can be tackled by gettig rid of the radial component. The Yogi \cite{yogi} utilizes the limit of variance in the stochastic gradients to control the learning rate. The AngularGrad \cite{AngularGrad} utilizes the change in gradient orientation to control the learning rate. In order to control the learning rate, decay based SGD approaches have been also exploited \cite{hsueh2019stochastic} \cite{kobayashi2021phase}.
Though the existing optimization methods try to control the learning rate by exploiting different properties of gradients, they still suffer due to inconsistent gradients. In this paper, we tackle this issue through the gradient norm correction to make it historically consistent throughout the training iterations.

\begin{algorithm}[!t]
\caption{Adam Optimizer}
\label{algo:adam}
\SetAlgoLined
\textbf{Initialize:} $\bm{\theta}_{0},\bm{m}_{0}\gets0,\bm{v}_{0}\gets0,t\gets0$\\
\textbf{Hyperparameters:} $\alpha, \beta_1, \beta_2$\\
\textbf{While} $\bm{\theta}_{t}$ not converged\\
    \hspace{0.45cm} $t \gets t+1$\\
    \hspace{0.45cm} $\bm{g}_t \gets \nabla_{\theta} f_t(\bm{\theta}_{t-1})$ \\
    \hspace{0.45cm} $\bm{m}_t \gets \beta_1\bm{m}_{t-1} + (1-\beta_1)\bm{g}_t$\\
    \hspace{0.45cm} $\bm{v}_t \gets \beta_2\bm{v}_{t-1} + (1-\beta_2)\bm{g}^2_t$\\
    \hspace{0.45cm} \textbf{Bias Correction}\\
    \hspace{0.9cm} $\widehat{\bm{m}_t} \gets \bm{m}_t/(1-\beta_1^t)$, $\widehat{\bm{v}_t} \gets \bm{v}_t/(1-\beta_2^t)$\\
    \hspace{0.45cm} \textbf{Update}\\
    \hspace{0.9cm} $\bm{\theta}_t \gets \bm{\theta}_{t-1} - \alpha \widehat{\bm{m}_t}/(\sqrt{\widehat{\bm{v}_t}} + \epsilon)$
\end{algorithm}

In this paper we tackle the above mentioned issues with the help of gradient norm correction by exploiting the history of gradient norm. The contributions are as follows:
\begin{enumerate}
    \item We propose an AdaNorm approach by exploiting the EMA of gradient norm of past iterations. The proposed AdaNorm rectifies the gradient norm based on the training history to better maintain the consistent and informative gradient.
    \item The proposed AdaNorm approach is generic and can be used with any existing adaptive SGD optimizer. We use the proposed AdaNorm with Adam \cite{adam}, diffGrad \cite{diffgrad}, Radam \cite{radam} and AdaBelief \cite{adabelief} optimizers and propose AdamNorm, diffGradNorm, RadamNorm and AdaBeliefNorm optimizers, respectively.
    \item We include an intuitive explanation and convergence proof for the proposed optimizer. We also show the impact of the proposed AdaNorm approach on the behaviour of gradient norm experimentally. 
    \item We perform a rigorous experimental study on three benchmark datasets, including CIFAR10, CIFAR100 and TinyImageNet for object recognition to demonstrate the efficacy of the proposed AdaNorm based optimizers. The impacts of hypermeter, AdaNorm on second moment, learning rate and bacth size are also studied in the experiments.
\end{enumerate}

We organize this paper by presenting the proposed AdaNorm optimizers in Section 2, Intuitive Explanation and Convergence Analysis in Section 3, Experimental settings in Section 4, Results \& discussion in Section 5, Ablation study in Section 6 and Conclusion in Section 7.

\begin{algorithm}[!t]
\caption{AdamNorm Optimizer}
\label{algo:adamnorm}
\SetAlgoLined
\textbf{Initialize:} $\bm{\theta}_{0},\bm{m}_{0}\gets0,\bm{v}_{0}\gets0,e_{0}\gets0,t\gets0$\\
\textbf{Hyperparameters:} $\alpha, \beta_1, \beta_2, \gamma$\\
\textbf{While} $\bm{\theta}_{t}$ not converged\\
    \hspace{0.45cm} $t \gets t+1$\\
    \hspace{0.45cm} $\bm{g}_t \gets \nabla_{\theta} f_t(\bm{\theta}_{t-1})$ \\
    \hspace{0.45cm} \textcolor{blue}{$g_{norm} \gets L_2Norm(\bm{g}_t)$}\\
    \hspace{0.45cm} \textcolor{blue}{$e_t = \gamma e_{t-1} + (1-\gamma)g_{norm}$}\\
    \hspace{0.45cm} \textcolor{blue}{$\bm{s}_t = \bm{g}_t$}\\
    \hspace{0.45cm} \textcolor{blue}{\textbf{If} $e_t > g_{norm}$}\\
    \hspace{0.9cm} \textcolor{blue}{$\bm{s}_t = (e_t / g_{norm})\bm{g}_t$}\\
    % \hspace{0.45cm} \textcolor{blue}{\textbf{Else}}\\
    % \hspace{0.90cm} \textcolor{blue}{$\bm{s}_t = \bm{g}_t$}\\
    \hspace{0.45cm} $\bm{m}_t \gets \beta_1 \bm{m}_{t-1} + (1-\beta_1) \textcolor{blue}{\bm{s}_t}$\\
    \hspace{0.45cm} $\bm{v}_t \gets \beta_2 \bm{v}_{t-1} + (1-\beta_2) \bm{g}^2_t$\\
    \hspace{0.45cm} \textbf{Bias Correction}\\
    \hspace{0.9cm} $\widehat{\bm{m}_t} \gets \bm{m}_t/(1-\beta_1^t)$, $\widehat{\bm{v}_t} \gets \bm{v}_t/(1-\beta_2^t)$\\
    \hspace{0.45cm} \textbf{Update}\\
    \hspace{0.9cm} $\bm{\theta}_t \gets \bm{\theta}_{t-1} - \alpha \widehat{\bm{m}_t}/(\sqrt{\widehat{\bm{v}_t}} + \epsilon)$
\end{algorithm}

\section{Proposed AdaNorm Optimizer}
Let consider a network ($f$) represented by its parameters $\bm{\theta}$ to be trained using SGD approach in an iterative manner. The parameters are initialized before start of the training and represented as $\bm{\theta}_0$. In any given $t^{th}$ iteration, the gradient of objective function w.r.t. the parameters (i.e., $\bm{g}_t$) is computed using chain-rule and expressed as,
\begin{equation}
\bm{g}_t = \nabla_{\theta} f_t(\bm{\theta}_{t-1})
\end{equation}
where $f_t$ is the model at $t^{th}$ iteration, $\bm{\theta}_{t-1}$ represent the parameters in the previous iteration, and $\nabla_{\theta}$ represent the gradient over parameters $\bm{\theta}$.

The existing optimizers, such as Adam (see Algorithm \ref{algo:adam}), diffGrad, Radam, and AdaBelief use the $\bm{g}_t$ to compute the Exponential Moving Average (EMA) of gradients ($\bm{m}_t$) which is used to update the parameters. However, the gradient $\bm{g}_t$ is computed as an average on a batch of randomly drawn training samples. Hence, it might not be representative, not consistent with the past gradient behaviour and prone to be bad. In this paper we tackle this problem by correcting the norm of the gradient of current batch with the help of historical gradient norm. 

Let $g_{norm} = ||\bm{g}_t||_2$ is the L2-Norm of the current gradient vector $\bm{g}_t = (g_{t,1}, g_{t,2}, ..., g_{t,k})$. The computation of the $g_{norm}$ can be given as,
\begin{equation}
    g_{norm} = \sqrt{\sum_{i=1}^{k}{(g_{t,i})^2}}
\end{equation}
where $g_{t,i}$ is the $i^{th}$ element of $\bm{g}_t$ and $k$ is the number of elements in $\bm{g}_t$.

Let represent the norm corrected gradient as $\bm{s}_t$. The computation of $\bm{s}_t$ is proposed as,
\begin{equation}
    \bm{s}_t = 
    \begin{cases}
    ({e_t}/{g_{norm}})\bm{g}_t,& \text{if } e_t > g_{norm}\\
    \bm{g}_t,  & \text{otherwise}
    \end{cases}
\end{equation}
where $e_t$ is the historical gradient norm computed in the $t^{th}$ iteration using the norm of past gradients, i.e., previous iterations. We use the EMA approach to compute $e_t$ as,
\begin{equation}
    e_t = \gamma e_{t-1} + (1-\gamma) g_{norm}
    \label{hyperparameter}
\end{equation}
where $g_{norm}$ is the L2-Norm of the current gradient $\bm{g}_t$, $e_{t-1}$ is the historical gradient norm computed in the previous $(t-1)^{th}$ training iteration with $e_0 = 0$ as the initial value before the start of the training and $\gamma$ is a hyperparameter to control the contribution of past historical gradient norm and current gradient norm in the computation of new historical gradient norm. The impact of $\gamma$ is analyzed in the experiments section. The proposed gradient norm correction step makes the norm of the current gradient to be at least the historical gradient norm. Inherently, it forces the current gradient to be better aligned and consistent with the behaviour of the previous gradients and tackles the problem of bad gradients in existing methods. Moreover, it reduces the dependency on batch size and makes the training of the deep network more effective and stable.

We integrate the proposed AdaNorm concept of gradient norm correction with Adam \cite{adam} and propose AdamNorm optimizer.
We use the gradient with corrected norm ($\bm{s}_t$) to compute the first moment $\bm{m}_t$ in the proposed AdamNorm optimizer, given as,
\begin{equation}
    \bm{m}_t = \beta_1\bm{m}_{t-1} + (1-\beta_1)\bm{s}_t
\end{equation}
where $\bm{m}_{t-1}$ and $\bm{m}_{t}$ are the first moment in $(t-1)^{th}$ and $t^{th}$ iterations, respectively, $\bm{m}_{0}$ is initialized with 0, and $\beta_1$ is a hyperparameter. However, we use the original gradient ($g_t$) to compute the second moment, given as,
\begin{equation}
    \bm{v}_t = \beta_2\bm{v}_{t-1} + (1-\beta_2)\bm{g}_t^2
    \label{secondmoment}
\end{equation}
where $\bm{v}_{t-1}$ and $\bm{v}_{t}$ are the second moment in $(t-1)^{th}$ and $t^{th}$ iterations, respectively, $\bm{v}_{0}$ is initialized with 0, and $\beta_2$ is a hyperparameter. As the second moment is used to control the learning rate, we avoid to use the norm corrected gradient in its computation as it may lead to significantly lower effective step-size and hamper the learning capability. The impact of gradient norm correction on second moment is analyzed in the experiments.

By following the Adam \cite{adam}, we perform the bias correction of moments to avoid very high step-size in the initial training iterations as follows,
\begin{equation}
    \widehat{\bm{m}_t} \gets \bm{m}_t/(1-\beta_1^t)    
\end{equation}
\begin{equation}
    \widehat{\bm{v}_t} \gets \bm{v}_t/(1-\beta_2^t)    
\end{equation}
where $t$ is the current iteration number, $\widehat{\bm{m}_t}$ and $\widehat{\bm{v}_t}$ are the first and second moment after bias correction, respectively.

Finally, the parameters of the network is updated based on $\widehat{\bm{m}_t}$ and $\widehat{\bm{v}_t}$ as follows,
\begin{equation}
    \bm{\theta}_t \gets \bm{\theta}_{t-1} - \alpha \widehat{\bm{m}_t}/(\sqrt{\widehat{\bm{v}_t}} + \epsilon)
\end{equation}
where $\bm{\theta}_{t-1}$ is the parameter after $(t-1)^{th}$ training iteration, $\bm{\theta}_{t}$ is the parameter after the current training iteration, and $\alpha$ is the learning rate used to compute the effective step-size for the parameter update. The steps of the proposed AdamNorm optimizer is summarized in Algorithm \ref{algo:adamnorm} with highlighted changes in Blue color w.r.t. the Adam.
    
Note that the proposed gradient norm correction using historical gradient norm is a generic idea and can be integrated with any existing SGD optimization technique. We just described above the steps of AdamNorm, i.e., the integration of the proposed concept with Adam \cite{adam}. However, in order to show the generalization of the gradient norm correction approach, we also integrate it with the recent state-of-the-art optimizers, including diffGrad \cite{diffgrad}, Radam \cite{radam} and AdaBelief \cite{adabelief} optimizers and propose diffGradNorm, RadamNorm and AdaBeliefNorm optimizers, respectively. The Algorithms of diffGrad, diffGradNorm, Radam, RadamNorm, AdaBelief and AdaBeliefNorm are provided in Supplementary.

\begin{figure}
    \centering
    \includegraphics[width=0.95\columnwidth]{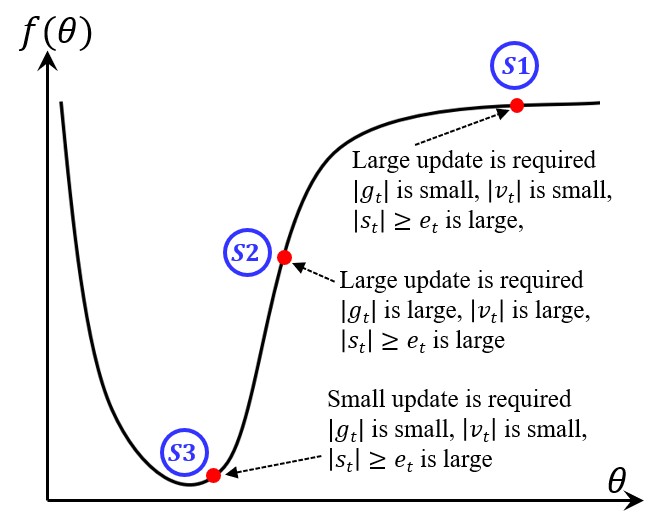}
    \caption{Typical scenarios depicting the importance of adaptive parameter update in optimization \cite{adabelief}.}
    \label{fig:justification}
\end{figure}

\section{Intuitive Explanation and Convergence Analysis}
\subsection{Intuitive Explanation}
In order to justify the importance of the gradient norm correction, we provide an intuitive explanation through Fig. \ref{fig:justification} that how the proposed AdaNorm approach provides a promising trade-off between large and small weight updates with the help of three typical scenarios in optimization on a one dimensional optimization curvature (i.e., S1, S2 and S3). The bias correction step is ignored for simplicity. The gradient norm $g_{norm}$ is considered as $|g_t|$ for one dimensional optimization.

The scenario \textbf{S1} depicts the flat region which is very common in optimization. In such region, an ideal optimizer expects the large update, but the gradient $g_t$ is very small. The small $g_t$ limits the $m_t$ in case of Adam leading to still small update. However, the $m_t$ is large in case of AdamNorm leading to large update as $s_t$ is large due to $|s_t| \geq e_t$ which is maintained to be sufficiently large historically over the training epochs. The $v_t$ is small and equally helpful in both the cases.

The scenario \textbf{S2} depicts the large gradient region in the optimization landscape. In such region, an ideal optimizer expects the large update, which is supported by large $g_t$. The $m_t$ in Adam is large in this case leading to large update. However, the $m_t$ in AdamNorm is at least the $m_t$ in Adam leading to large update. It shows that AdamNorm can perform at least as good as Adam in large gradient region. The effect of $v_t$ is similar in both the cases. 

The scenario \textbf{S3} depicts the steep and narrow valley region in the optimization landscape, which mimics the minimum of function. In such region, an ideal optimizer expects the small update, which is supported by small $g_t$. The Adam leads to small $m_t$ leading to small update, but the AdamNorm leads to relatively large $m_t$ leading to relatively large update which might be favourable to unwanted local minimum. In case of minimum, the $m_t$ in AdamNorm will become small in few iterations of parameter updates near minimum which will eventually lead to convergence. The $v_t$ behaves equally bad in both the cases.

\begin{table*}[!t]
    \caption{Classification results in terms of accuracy (\%) on CIFAR10, CIFAR100 and TinyImageNet datasets using Adam, diffGrad, Radam and AdaBelief without and with the proposed AdaNorm technique. The value of $\gamma$ is set to 0.95 in this experiment and results are computed as an average over three independent runs.}
    \centering
    % \resizebox{\textwidth}{!}{%
    \begin{tabular}{p{0.075\textwidth}|p{0.057\textwidth}p{0.122\textwidth}|p{0.057\textwidth}p{0.112\textwidth}|p{0.057\textwidth}p{0.112\textwidth}|p{0.065\textwidth}p{0.12\textwidth}}
        \hline
        & \multicolumn{8}{c}{Classification accuracy (\%) using different optimizers without and with AdaNorm} \\
        \cline{2-9}
        CNN & \multicolumn{2}{c|}{Adam} & \multicolumn{2}{c|}{diffGrad} & \multicolumn{2}{c|}{Radam} & \multicolumn{2}{c}{AdaBelief}\\
        \cline{2-9}
        Models & Adam & AdamNorm & diffGrad & diffGradNorm & Radam & RadamNorm & AdaBelief & AdaBeliefNorm\\
        \hline
         
        \multicolumn{9}{c}{Results on CIFAR10 Dataset}\\
        \hline
        VGG16 & 92.55 &	\textbf{92.83}	($\uparrow$	0.30) &	92.76 &	\textbf{92.87} ($\uparrow$ 0.12) &	92.94 & \textbf{93.14} ($\uparrow$	0.22) &	92.71 &	\textbf{92.81} ($\uparrow$	0.11) \\
        ResNet18 & 93.54 & \textbf{93.78} ($\uparrow$	0.26) & 93.49 & \textbf{93.98} ($\uparrow$	0.52) & 93.82 & \textbf{93.89} ($\uparrow$	0.07) & 93.63 & \textbf{93.66} ($\uparrow$ 0.03) \\
        ResNet50 & 93.83 & \textbf{94.01} ($\uparrow$	0.19) & 93.81 & \textbf{94.23} ($\uparrow$	0.45) & 94.14 & \textbf{94.21} ($\uparrow$	0.07) & 94.1 & \textbf{94.16} ($\uparrow$ 0.06) \\        
        \hline
         
        \multicolumn{9}{c}{Results on CIFAR100 Dataset}\\
        \hline
        VGG16 & 67.29 &	\textbf{69.15}	($\uparrow$	2.76) &	68.19 &	\textbf{68.31} ($\uparrow$ 0.18) &	70.69 & \textbf{70.77} ($\uparrow$	0.11) &	68.92 &	\textbf{69.24}	($\uparrow$	0.46) \\
        ResNet18 & 71.09 & \textbf{73.11} ($\uparrow$	2.84) & 73.5 & \textbf{73.64} ($\uparrow$	0.19) & 73.22 & \textbf{73.34} ($\uparrow$	0.16) & 72.72 & \textbf{73.31} ($\uparrow$ 0.81) \\
        ResNet50 & 71.88 & \textbf{75.53} ($\uparrow$	5.08) & 75.06 & \textbf{75.49} ($\uparrow$	0.57) & 74.95 & \textbf{75.39} ($\uparrow$	0.59) & \textbf{75.53} & 75.49 ($\downarrow$	0.05) \\        
        \hline
         
        \multicolumn{9}{c}{Results on TinyImageNet Dataset}\\
        \hline
        VGG16 & 41.93 &	\textbf{44.67}	($\uparrow$	6.53) &	42.91 &	\textbf{43.49} ($\uparrow$ 1.35) &	43.84 & \textbf{45.02} ($\uparrow$	2.69) &	44.23 &	\textbf{44.79}	($\uparrow$	1.27) \\
        ResNet18 & 47.73 & \textbf{49.57} ($\uparrow$	3.86) & 49.34 & \textbf{49.80} ($\uparrow$	0.93) & 48.73 & \textbf{50.50} ($\uparrow$	3.63) & 49.25 & \textbf{49.99} ($\uparrow$	1.50) \\
        ResNet50 & 48.98 & \textbf{54.44} ($\uparrow$	11.15) & 51.32 & \textbf{53.75} ($\uparrow$	4.73) & 51.63 & \textbf{52.87} ($\uparrow$	2.40) & 53.57 & \textbf{54.44} ($\uparrow$	1.62) \\        
         \hline
    \end{tabular}
    % }
\label{tab:results_comparison}
\end{table*}

\subsection{Convergence Analysis}
We use the online learning framework proposed in \cite{zinkevich2003online} to show the convergence property of AdamNorm similar to Adam \cite{adam}. Assume $f_1(\theta)$, $f_2(\theta)$,$...$, $f_T(\theta)$ as the convex cost functions in an unknown sequence. We compute the regret bound as follows,
\begin{equation}
R(T)=\sum_{t=1}^{T}{[f_t(\theta_t)-f_t(\theta^*)]}
\end{equation}
where $f_t(\theta_t)$ is the $t^{th}$ online guess, $f_t(\theta^*)$ is the best parameter setting from a feasible set $\chi$ with $\theta^*=\mbox{arg }\mbox{min}_{\theta \in \chi }\sum_{t=1}^{T}{f_t(\theta)}$. It is observed that the regret bound of AdamNorm is similar to Adam, i.e., $O(\sqrt{T})$. We provide the convergence proof of the AdamNorm in Supplementary. Let $g_{t,i}$ and $s_{t,i}$ are the gradient and the norm rectified gradient, respectively, in the $t^{th}$ iteration for the $i^{th}$ element, $g_{1:t,i}=[g_{1,i},g_{2,i},...,g_{t,i}] \in \mathbb{R}^t$ and $s_{1:t,i}=[s_{1,i},s_{2,i},...,s_{t,i}] \in \mathbb{R}^t$ are the gradient vector and the norm rectified gradient vector, respectively, for the $i^{th}$ parameter over all iterations up to $t$, and $\eta \triangleq \frac{\beta_1^2}{\sqrt{\beta_2}}$. 

\begin{theorem}
\textit{Let the gradients for function $f_t$ are bounded (i.e., $||g_{t,\theta}||_2 \leq G$ and $||g_{t,\theta}||_{\infty} \leq G_{\infty}$ for all $\theta \in R^d$). Let the distance produced by AdamNorm between any $\theta_t$ are also bounded (i.e., $||\theta_n-\theta_m||_2 \leq D$ and $||\theta_n-\theta_m||_\infty \leq D_\infty$ for any $m,n\in\{1,...,T\}$). Let $\eta \triangleq \frac{\beta_1^2}{\sqrt{\beta_2}}$, $\beta_1,\beta_2 \in [0,1)$ satisfy $\frac{\beta_1^2}{\sqrt{\beta_2}} < 1$, $\alpha_t=\frac{\alpha}{\sqrt{t}}$, and $\beta_{1,t}=\beta_1\lambda^{t-1},\lambda \in (0,1)$ where $\lambda$ is typically very close to $1$, e.g., $1-10^{-8}$. For all $T \geq 1$, the AdamNorm shows the following guarantee:}
\begin{equation*}
\begin{split}
R(T) & \leq \frac{D^2}{2\alpha(1-\beta_1)}\sum_{i=1}^{d}{\sqrt{T\hat{v}_{T,i}}} 
\\&+ \frac{\alpha(1+\beta_1) G^3_\infty}{(1-\beta_1)\sqrt{1-\beta_2}(1-\gamma)^2G^2}\sum_{i=1}^{d}{||g_{1:T,i}||_2} 
\\&+ \sum_{i=1}^{d}{\frac{D_{\infty}^{2}G_{\infty}\sqrt{1-\beta_2}}{2\alpha (1-\beta_1)(1-\lambda)^2}}
\end{split}
\end{equation*}
\end{theorem}
Note that the additive term over the dimension ($d$) can be much smaller than its upper bound $\sum_{i=1}^{d}{||g_{1:T,i}||_2}<< dG_\infty\sqrt{T}$ and $\sum_{i=1}^{d}{\sqrt{T\hat{v}_{T,i}}} << dG_\infty\sqrt{T}$. In general, $O(\log d\sqrt{T})$ is achieved by adaptive methods such as Adam and AdamNorm which is better than the $O(\sqrt{dT})$ of non-adaptive methods. The following corollary depicts the convergence of average regret of AdamNorm.
\begin{corollary}
\textit{Consider the bounded gradients for function $f_t$ (i.e., $||g_{t,\theta}||_2 \leq G$ and $||g_{t,\theta}||_{\infty} \leq G_{\infty}$) for all $\theta \in R^d$. Also, assume that the AdamNorm produces the bounded distance between any $\theta_t$ (i.e., $||\theta_n-\theta_m||_2 \leq D$ and $||\theta_n-\theta_m||_\infty \leq D_\infty$ for any $m,n\in\{1,...,T\}$). For all $T \geq 1$, the proposed AdamNorm optimizer shows the following guarantee:} 
\begin{equation*}
\frac{R(T)}{T}=O(\frac{1}{\sqrt{T}}). 
\end{equation*}
Thus, $\lim_{T\rightarrow\infty}\frac{R(T)}{T}=0$.
\end{corollary}

Theoretically, the convergence rate in terms of regret bounds for AdamNorm is similar to Adam-type optimizers (i.e., $O(\sqrt{T})$) \cite{adam}, \cite{diffgrad}, \cite{radam}, \cite{adabelief}, which is computed in the worst possible case. However, the empirical analysis suggests that the AdamNorm outperforms Adam mainly because the cases as detailed in Section 3.1, which occur more frequently.
% It is possible that the above bounds are loose due to our similar or weaker assumptions. 

\section{Experimental Settings}
This section provides the details of CNN models used, datasets used and training settings.

% \subsection{CNN Models Used}
\textbf{CNN Models Used:}
In order to validate the efficacy of the proposed optimizers three CNN models, including VGG16 \cite{vgg}, ResNet18 and ResNet50 \cite{resnet}, are used in the experiments. The VGG16 is a simple CNN model, whereas the ResNet18 and ResNet50 are the residual connection based CNN models. The ResNet50 is a deep CNN model as compared to the VGG16 and ResNet18.

% \subsection{Datasets Used}
\textbf{Datasets Used:}
We validate the performance of the proposed optimizers on three standard visual recognition datasets, including CIFAR10 \cite{cifar}, CIFAR100 \cite{cifar}, and TinyImageNet \cite{tinyimagenet}. The CIFAR10 and CIFAR100 datasets contain 50000 images for training and 10000 images for testing. The CIFAR10 contains 10 object classes with equal number of samples. However, the CIFAR100 contains 100 object classes with equal number of samples. The CIFAR100 is a fine-grained dataset. The TinyImageNet dataset contains 200 object classes with 500 training images per class (i.e., total 100000 training images) and 50 test images per class (i.e., total 10000 test images).

% \subsection{Training Settings}
\textbf{Training Settings:}
We perform the experiments using the Pytorch framework and train all the CNN models using Google Colab based freely available computational resources with single GPU. The training is performed for 100 Epochs with a batch size of 64. The learning rate is set to 0.001 initially and dropped to 0.0001 after 80 Epoch of training. For a fair comparison we consider the same common hyperparameters for all the optimizers, i.e., $\beta_1 = 0.9$ and $\beta_2 = 0.999$.
The training and test images are normalized as per the standard practice. 
The data augmentations with random cropping, random horizontal flipping and normalization with mean (0.4914, 0.4822, 0.4465) and standard deviation (0.2023, 0.1994, 0.2010) are performed during training. Only normalization is used during testing.

\begin{figure*}[t]
    \centering
    \includegraphics[trim={2mm 0.5mm 2mm 2mm},clip, width=0.328\textwidth, height=3.75cm]{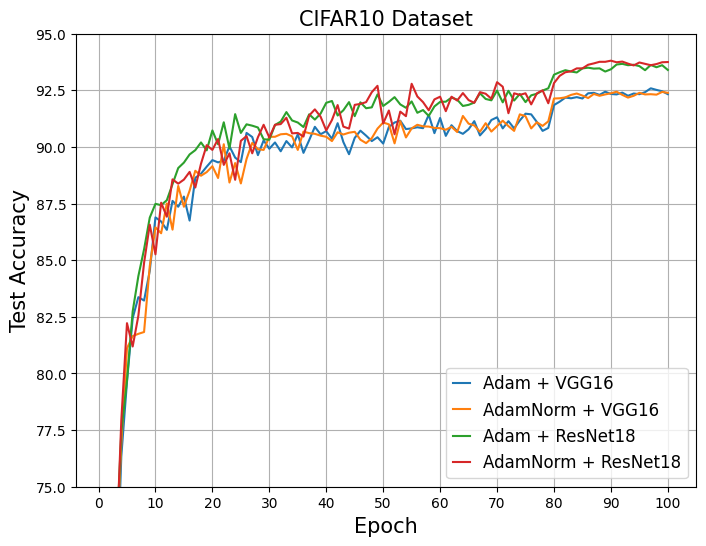}
    \includegraphics[trim={2mm 0.5mm 2mm 2mm},clip, width=0.328\textwidth, height=3.75cm]{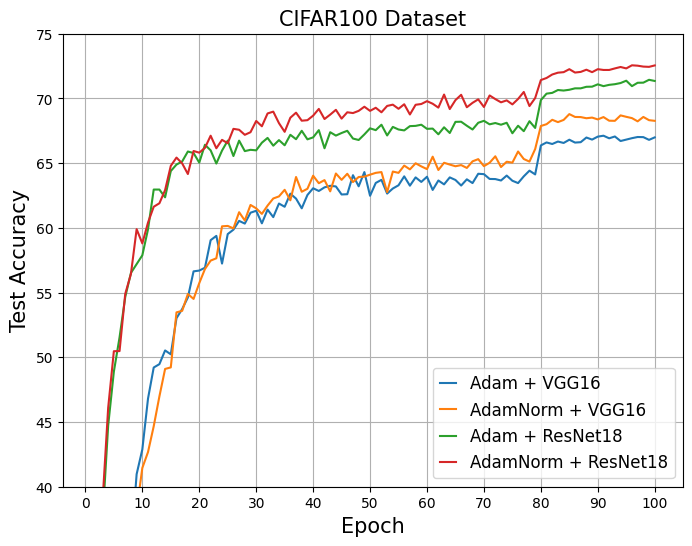}
    \includegraphics[trim={2mm 0.5mm 2mm 2mm},clip, width=0.328\textwidth, height=3.75cm]{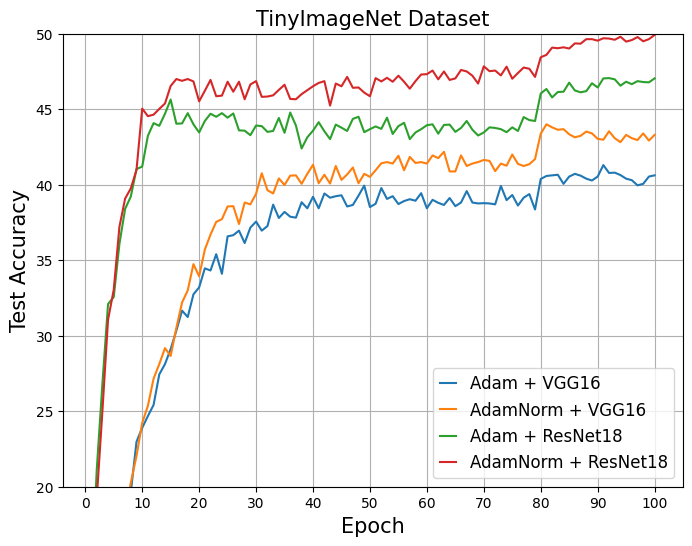}
    \includegraphics[trim={2mm 0.5mm 2mm 2mm},clip, width=0.328\textwidth, height=3.75cm]{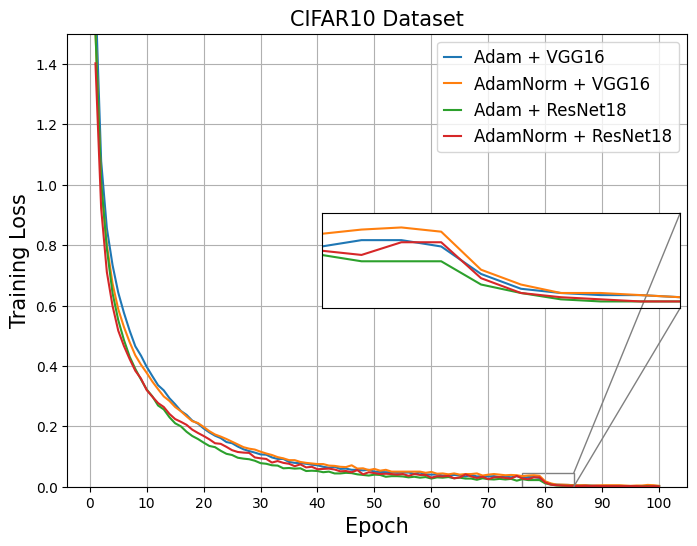}
    \includegraphics[trim={2mm 0.5mm 2mm 2mm},clip, width=0.328\textwidth, height=3.75cm]{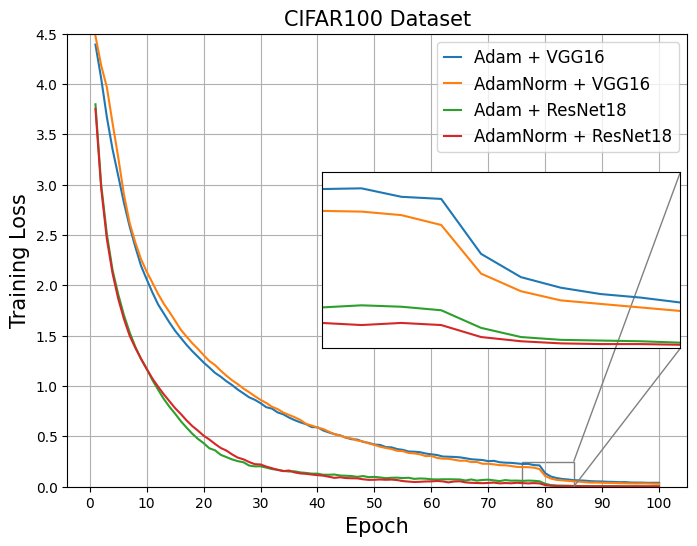}
    \includegraphics[trim={2mm 0.5mm 2mm 2mm},clip, width=0.328\textwidth, height=3.75cm]{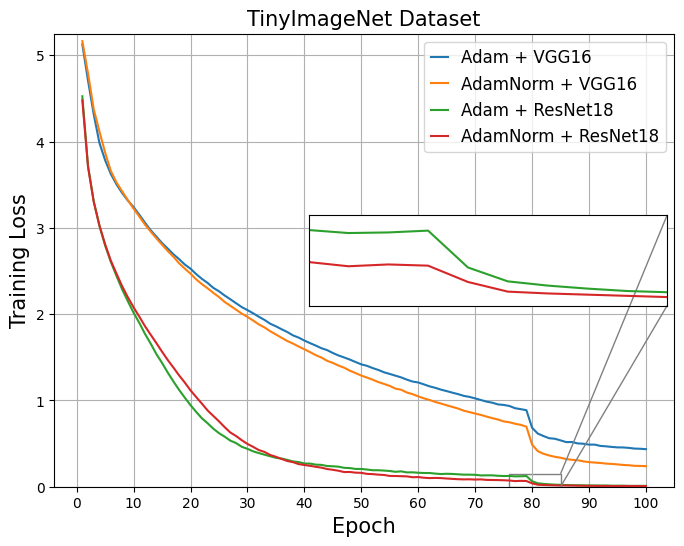}
    \caption{Test Accuracy (top row) and Training Loss (bottom row) vs Epoch plots using the Adam and AdamNorm optimizers for VGG16 and ResNet18 models on CIFAR10 (left), CIFAR100 (middle) and TinyImageNet (right) datasets. The value of $\gamma$ is $0.95$ in AdamNorm in this experiment. (Best viewed in color) }
    \label{fig:loss}
\end{figure*}

\begin{figure*}[t]
    \centering
    \includegraphics[trim={4mm 1mm 8mm 7mm},clip, width=0.246\textwidth]{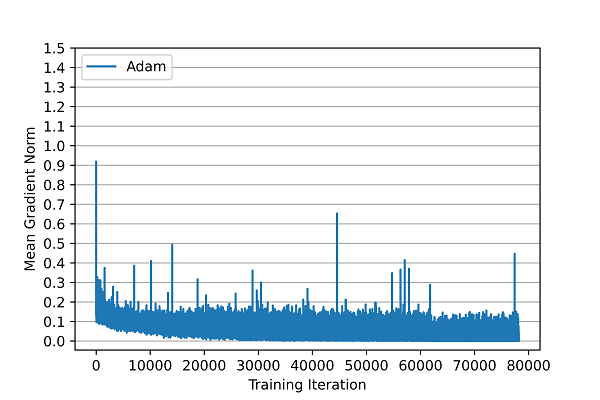}
    \includegraphics[trim={4mm 1mm 8mm 7mm},clip, width=0.246\textwidth]{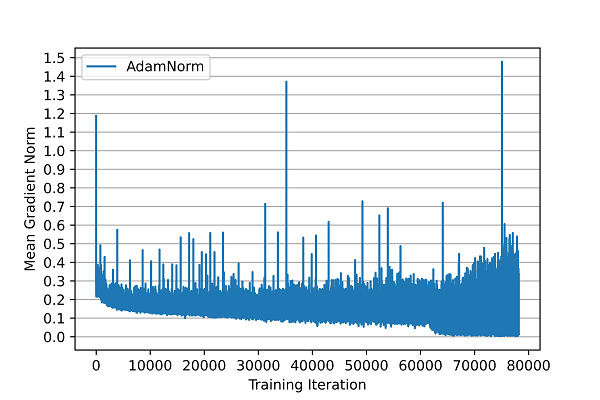}
    \includegraphics[trim={4mm 1mm 8mm 7mm},clip, width=0.246\textwidth]{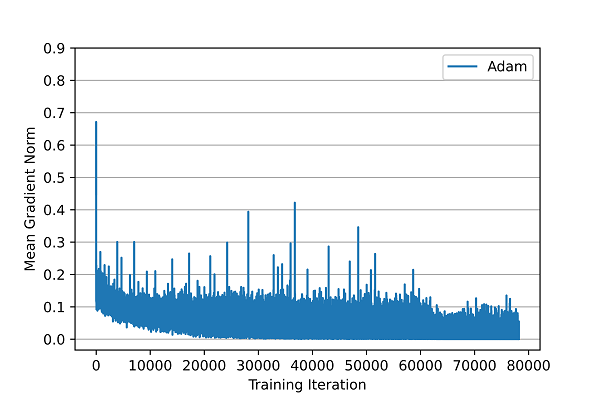}
    \includegraphics[trim={4mm 1mm 8mm 7mm},clip, width=0.246\textwidth]{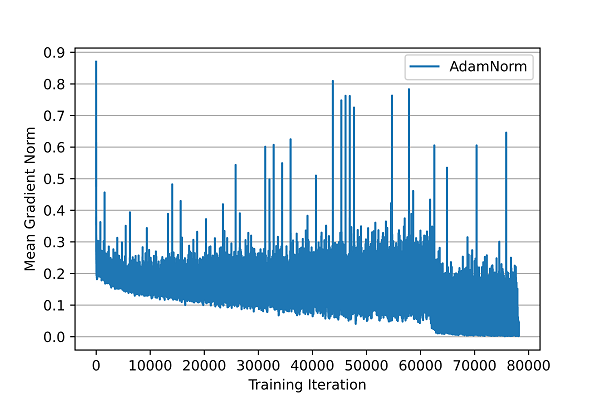}
    \includegraphics[trim={4mm 1mm 8mm 7mm},clip, width=0.246\textwidth]{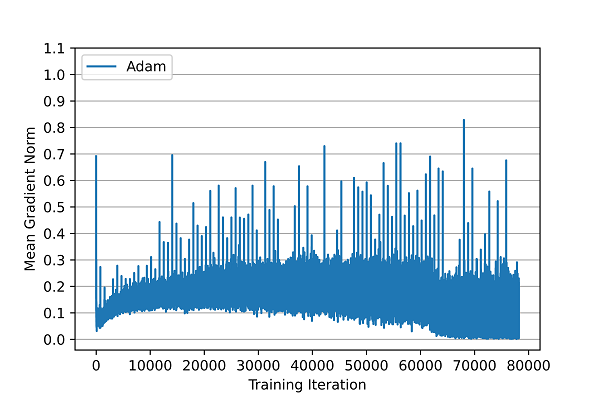}
    \includegraphics[trim={4mm 1mm 8mm 7mm},clip, width=0.246\textwidth]{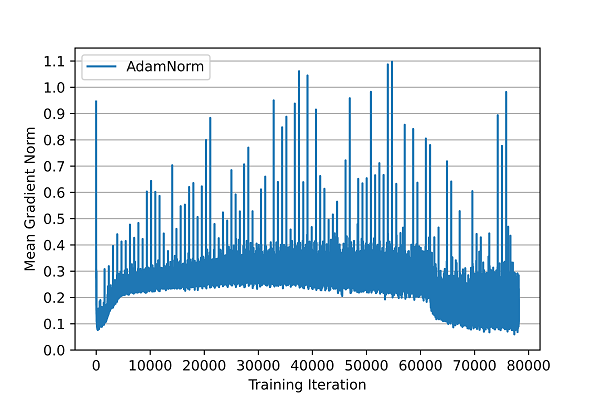}
    \includegraphics[trim={4mm 1mm 8mm 7mm},clip, width=0.246\textwidth]{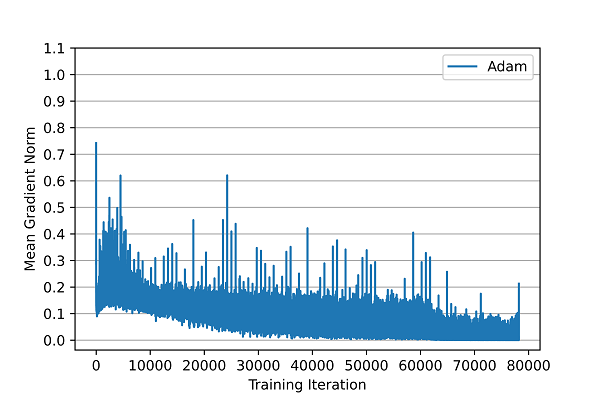}
    \includegraphics[trim={4mm 1mm 8mm 7mm},clip, width=0.246\textwidth]{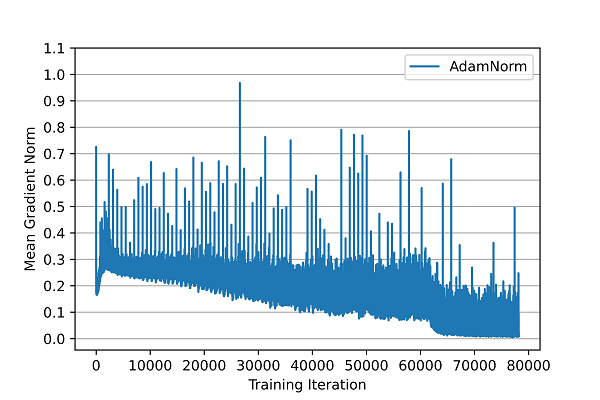}
    \includegraphics[trim={4mm 1mm 8mm 7mm},clip, width=0.246\textwidth]{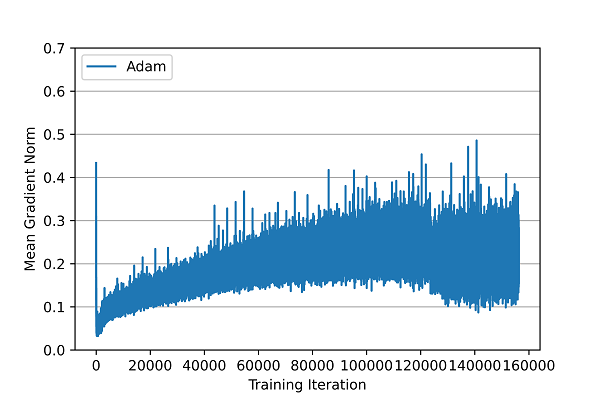}
    \includegraphics[trim={4mm 1mm 8mm 7mm},clip, width=0.246\textwidth]{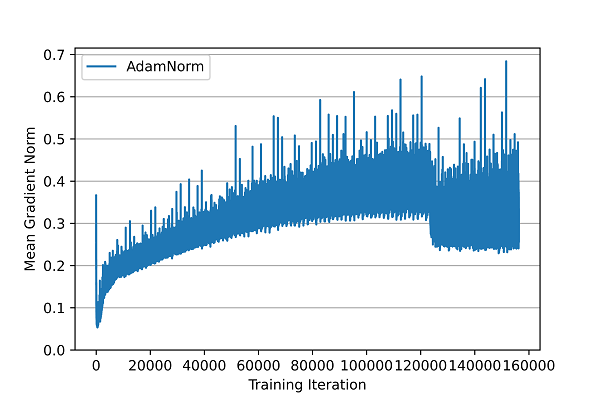}
    \includegraphics[trim={4mm 1mm 8mm 7mm},clip, width=0.246\textwidth]{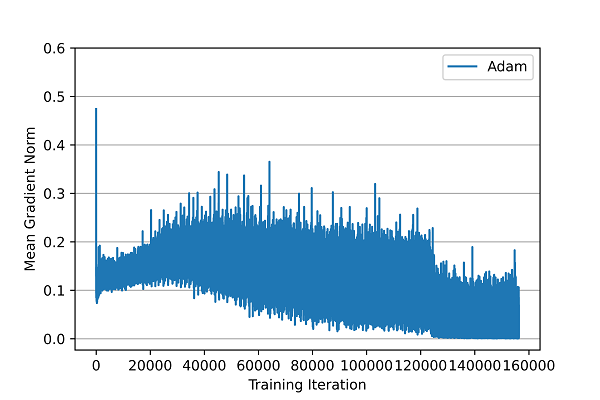}
    \includegraphics[trim={4mm 1mm 8mm 7mm},clip, width=0.246\textwidth]{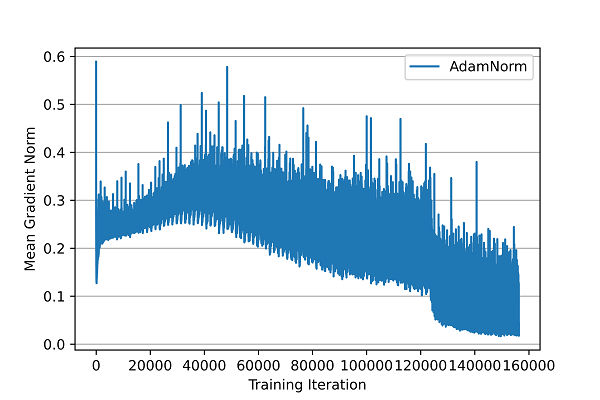}    
    \caption{The Mean Gradient Norm using the Adam and AdamNorm optimizers (i.e., without and with the proposed history based gradient norm correction, respectively) at different training iterations. The plots in first, second and third row correspond to CIFAR10, CIFAR100 and TinyImageNet datasets, respectively. The plots in first and second column are computed using VGG16 and the plots in third and fourth column are computed using ResNet18. The gradient norm using the optimizers with proposed method is significantly greater than the corresponding vanilla optimizers. Note that the change in gradient norm at Epoch 80 is due to drop in the learning rate.}
    \label{fig:gradient_norm}
\end{figure*}

\begin{table*}[t]
    \caption{The effect of hyperparameter $\gamma$ used in the EMA of the proposed history based gradient norm computation. The results are computed as an average over three independent runs using AdamNorm optimizer.}
    \centering
    \begin{tabular}{p{0.13\textwidth}|p{0.065\textwidth}p{0.065\textwidth}p{0.075\textwidth}|p{0.065\textwidth}p{0.065\textwidth}p{0.075\textwidth}|p{0.065\textwidth}p{0.065\textwidth}p{0.075\textwidth}}
      \hline
      Hyperparameter & \multicolumn{3}{c|}{CIFAR10 Dataset} & \multicolumn{3}{c|}{CIFAR100 Dataset} & \multicolumn{3}{c}{TinyImageNet Dataset}\\ \cline{2-10}
      ($\gamma$) & VGG16 & ResNet18 & ResNet50  & VGG16 & ResNet18 & ResNet50 & VGG16 & ResNet18 & ResNet50 \\\hline
      0.9 & 92.56 & 93.72 & 93.95 & 68.85 & 73.06 & \textbf{75.53} & 44.85 & \textbf{50.47} & \textbf{54.72}\\
      0.95 & 92.83 & \textbf{93.78} & 94.01 & 69.15 & 73.11 & \textbf{75.53} & 44.67 & 49.57 & 54.44\\
      0.99 & \textbf{92.88} & 93.55 & \textbf{94.11} & \textbf{69.33} & \textbf{73.13} & 75.43 & \textbf{45.08} & 49.41 & 54.54\\
      0.999 & 92.66 & 93.68 & 93.54 & 69.18 & 73.16 & 75.48 & 44.71 & 50.45 & 54.14\\
      \hline
    \end{tabular}
    \label{tab:hyperparameter}
\end{table*}

\begin{table*}[t]
    \caption{The results by applying the normalized gradient in different combination of first and second order moment. The value of $\gamma$ is set to 0.95 in this experiment and results are computed as an average over three independent runs.}
    \centering
    \begin{tabular}{p{0.132\textwidth}|p{0.065\textwidth}p{0.065\textwidth}p{0.075\textwidth}|p{0.065\textwidth}p{0.065\textwidth}p{0.075\textwidth}|p{0.065\textwidth}p{0.065\textwidth}p{0.075\textwidth}}
      \hline
      Normalized & \multicolumn{3}{c|}{CIFAR10 Dataset} & \multicolumn{3}{c|}{CIFAR100 Dataset} & \multicolumn{3}{c}{TinyImageNet Dataset}\\ \cline{2-10}
      Gradient Setting & VGG16 & ResNet18 & ResNet50  & VGG16 & ResNet18 & ResNet50 & VGG16 & ResNet18 & ResNet50 \\\hline
      $1^{st}$ Moment & \textbf{92.83} & \textbf{93.78} & \textbf{94.01} & \textbf{69.15} & \textbf{73.11} & \textbf{75.53} & \textbf{44.67} & \textbf{49.57} & \textbf{54.44}\\
      $2^{nd}$ Moment & 92.74 & 93.67 & 93.36 & 68.86 & 72.65 & 74.30 & 43.43 & 49.17 & 54.01\\
      Both Moments & 92.69 & 93.61 & 93.10 & 69.08 & 72.95 & 74.54 & 43.77 & 49.19 & 53.73\\
      \hline
    \end{tabular}
    \label{tab:moment}
\end{table*}

\section{Experimental Results and Discussion}
This section provides the results comparison, experimental convergence analysis and the impact of the AdaNorm on the norm of the gradients. 

\subsection{Results Comparison}
The results comparison of the proposed gradient norm correction based optimizers are presented in Table \ref{tab:results_comparison} in terms of the accuracy (\%). We use four state-of-the-art adaptive optimizers (i.e., Adam \cite{adam}, diffGrad \cite{diffgrad}, Radam \cite{radam} and AdaBelief \cite{adabelief}) for the results comparison by applying the proposed concept with these optimizers. The results are compared using VGG16, ResNet18 and ResNet50 models on CIFAR10, CIFAR100 and TinyImageNet datasets. The value of $\gamma$ is set to 0.95 in this experiment and results are computed as an average over three independent runs. The higher results for an optimizer is highlighted in bold. The improvement in $\%$ due to the propsoed gradient norm correction is also highlighted in Table \ref{tab:results_comparison} with $\uparrow$ symbol. It can noticed that the accuracy due to the proposed optimizers is improved in almost all the cases. The performance is significantly improved on TinyImageNet dataset with highest improvement of 11.15\% using AdamNorm optimizer as compared to Adam for ResNet50 model. The consistent improvement in the performance using different CNN models and optimizers confirm the importance of the gradient norm correction based on the history of the gradient norm.

\subsection{Experimental Convergence Analysis}
In order to highlight the improved convergence due to the proposed gradient norm correction based optimizer, we plot the test accuracy (top row) and training loss (bottom row) obtained at every epoch using Adam and AdamNorm (with $\gamma=0.95$) optimizers for VGG16 and ResNet18 models on CIFAR10, CIFAR100 and TinyImageNet datasets in Fig. \ref{fig:loss}. The test accuracy plots depict that the performance of the proposed AdamNorm is consistently better than the Adam on CIFAR100 and TinyImageNet datasets and slightly better on CIFAR10 dataset. The training loss curve for the AdamNorm is also better than the Adam on CIFAR100 and TinyImageNet dataset, while it is comparable on CIFAR10 dataset. From the training loss plots on CIFAR100 and TinyImageNet datasets, it is clear that the Adam optimizer initially converges faster, but get saturated soon due to the lack of consistent gradients over the training epochs. However, the proposed AdamNorm optimizer makes the consistent updates due to the norm corrected gradients used for updates and leads to significantly lower loss as compared to the Adam optimizer. It confirms the need of history based gradient norm correction for better optimization of CNNs.

\subsection{Impact of Proposed Gradient Norm Correction}
The proposed gradient norm correction aims to enforce the gradient norm at any training iteration to follow the trend of gradient norms of past training iterations. In order to observe the impact of the gradient norm correction, we plot the mean gradient norm of the Adam and AdamNorm at each training iteration in Fig. \ref{fig:gradient_norm} using VGG16 ($1^{st}$ and $2^{nd}$ columns) and ResNet18 ($3^{rd}$ and $4^{th}$ columns) models on CIFAR10, CIFAR100 and TinyImageNet datasets, in $1^{st}$, $2^{nd}$ and $3^{rd}$ row, respectively. It can be seen that the gradient norm of the AdamNorm is much higher and representative than the Adam in all the cases. It can also be observed that the gradient norm of the AdamNorm is better aligned with the historical trend set by the past training iterations. The improved representation of the gradient norm is the key to the performance improvement of the proposed AdaNorm based optimizers.

\section{Ablation Study}
This ablation study presents the effect of AdaNorm hyperparameter, second moment, learning rate \& batch size.

\subsection{Impact of AdaNorm Hyperparameter}
In the proposed approach, the history of gradient norm is accumulated using EMA of gradient norms using a hyperparameter ($\gamma$) in (\ref{hyperparameter}). In the results comparison we use the value of $\gamma$ as $0.95$. However, in this experiment, we compute the results using the proposed AdamNorm optimizer for $\gamma = \{0.9, 0.95, 0.99, 0.999\}$ using the VGG16, ResNet18 and ResNet50 models on the CIFAR10, CIFAR100 and TinyImageNet datasets and report in Table \ref{tab:hyperparameter}. The results suggest that the value of $\gamma$ is not recommended to be very high such as $0.999$. Overall, relatively higher $\gamma$ such as $0.99$ is better suitable for the CIFAR10 and CIFAR100 datasets. However, relatively lower $\gamma$ such as $0.90$ leads to better performance on the TinyImageNet dataset. This behaviour is justified from the fact that the number of training iterations on the TinyImageNet is much higher as compared to the CIFAR10 and CIFAR100 datasets.

\subsection{Impact of AdaNorm on Second Moment}
In the proposed approach, the gradient norm correction is only applied on the first moment. In this experiment, we compute the results by using the gradient norm correction in second moment also (i.e., using $\bm{s}_t^2$ instead of $\bm{g}_t^2$ in (\ref{secondmoment})). Basically, we compute the results using the proposed AdamNorm optimizer by applying the gradient norm correction in three settings, a) in $1^{st}$ moment only, b) in $2^{nd}$ moment only, and c) in both $1^{st}$ and $2^{nd}$ moments. The results are summarized for VGG16, ResNet18 and ResNet50 models on CIFAR10, CIFAR100 and TinyImageNet datasets in Table \ref{tab:moment}. It is evident that the performance of the proposed optimizer is best when the gradient norm correction is applied only on the $1^{st}$ moment, because the $2^{nd}$ moment controls the learning rate and applying the gradient norm correction on the $2^{nd}$ moment hampers the effective step-size leading to poor performance.

\begin{table}[t]
    \caption{The impact of learning rate ($\alpha$). After 80 training Epochs, $\alpha$ is divided by 10. The $\gamma$ is set to 0.95 in this experiment and results are computed as an average over three independent runs.}
    \centering
    \begin{tabular}{p{0.07\textwidth}|p{0.035\textwidth}p{0.035\textwidth}p{0.044\textwidth}|p{0.035\textwidth}p{0.035\textwidth}p{0.038\textwidth}}
    \hline
    Model & \multicolumn{3}{c|}{Adam Optimizer} & \multicolumn{3}{c}{AdamNorm Optimizer}\\ \hline
    $\alpha$ & 0.01 & 0.001 & 0.0001 & 0.01 & 0.001 & 0.0001 \\\hline
    \multicolumn{6}{c}{CIFAR10 Dataset}\\\hline
    VGG16 & 92.21 & \textbf{92.55} & 92.02 & 92.30 & \textbf{92.83} & 92.27 \\
    ResNet18 & 92.90 & \textbf{93.54} & 93.07 & 93.18 & \textbf{93.78} & 93.13 \\\hline
    \multicolumn{6}{c}{CIFAR100 Dataset}\\\hline
    VGG16 & 65.93 & 67.29 & \textbf{68.57} & 66.19 & \textbf{69.15} & 68.75 \\
    ResNet18 & 66.84 & 71.09 & \textbf{72.95} & 68.42 & 73.11 & \textbf{73.14} \\\hline
    \multicolumn{6}{c}{TinyImageNet Dataset}\\\hline
    VGG16 & 36.78 & 41.93 & \textbf{44.55} & 41.41 & 44.67 & \textbf{46.65} \\
    ResNet18 & 45.50 & 47.73 & \textbf{48.87} & 46.94 & 49.57 & \textbf{51.08}\\
    \hline
    \end{tabular}
    \label{tab:lr}
\end{table}

\subsection{Impact of Learning Rate}
We also study the impact of learning rate ($\alpha$) on the proposed AdamNorm optimizer. The classification accuracies are summarized in Table \ref{tab:lr} on CIFAR10, CIFAR100 and TinyImageNet datasets using VGG16 and ResNet18 models for Adam and AdamNorm optimizers under different settings of learning rate, i.e., $\alpha = {0.01, 0.001, 0.0001}$. Note that the learning rate is divided by 10 after Epoch no. 80 in all the experiments. The results are reported as an average over three runs. The value of $\gamma$ is set to $0.95$. It is observed that the performance of AdamNorm is always better than Adam with same learning rate schedule. The results suggest that the smaller learning rate is better suitable on TinyImageNet datasets. However, the original considered learning rate (i.e., $0.001$) is reasonable on CIFAR10 and CIFAR100 datasets using the AdamNorm optimizer.

\begin{table}[t]
    \caption{The impact of batch size ($BS$). The $\alpha$ is $0.001$ the $\gamma$ is set to 0.95 in this experiment. Results are computed as an average over three independent runs.}
    \centering
    \begin{tabular}{p{0.07\textwidth}|p{0.037\textwidth}p{0.037\textwidth}p{0.037\textwidth}|p{0.037\textwidth}p{0.037\textwidth}p{0.037\textwidth}}
    \hline
    Model & \multicolumn{3}{c|}{Adam Optimizer} & \multicolumn{3}{c}{AdamNorm Optimizer}\\ \hline
    BS & 32 & 64 & 128 & 32 & 64 & 128 \\\hline
    \multicolumn{6}{c}{CIFAR10 Dataset}\\\hline
    VGG16 & \textbf{92.56} & 92.55 & 92.33 & 92.80 & \textbf{92.83} & 92.57 \\
    ResNet18 & 93.45 & 93.54 & \textbf{93.60} & 93.51 & \textbf{93.78} & 93.69 \\\hline
    \multicolumn{6}{c}{CIFAR100 Dataset}\\\hline
    VGG16 & 67.58 & 67.29 & \textbf{67.80} & 68.81 & 69.15 & \textbf{69.45} \\
    ResNet18 & 70.13 & 71.09 & \textbf{71.62} & \textbf{73.43} & 73.11 & 72.90 \\\hline
    \multicolumn{6}{c}{TinyImageNet Dataset}\\\hline
    VGG16 & \textbf{42.11} & 41.93 & 41.77 & 42.83 & \textbf{44.67} & 43.25 \\
    ResNet18 & 46.77 & 47.73 & \textbf{48.31} & 49.38 & 49.57 & \textbf{50.59} \\
    \hline
    \end{tabular}
    \label{tab:bs}
\end{table}

\subsection{Impact of Batch Size}
We also report the results by considering different batch sizes ($BS$), such as $32$, $64$, and $128$, in Table \ref{tab:bs} for the Adam and AdamNorm optimizers on CIFAR10, CIFAR100 and TinyImageNet datasets using VGG16 and ResNet18 models. The results are computed as an average over three runs. The values of $\alpha$ and $\gamma$ are $0.001$ and $0.95$, respectively. It is observed that the performance of the proposed AdamNorm optimizer is always better than the performance of Adam for all the batch sizes. The results of Adam are mostly better with large batch size. However, the results of AdamNorm are better for all the batch sizes in some or other cases. It shows that the proposed optimizer is more robust to the batch size, because the norm of the gradients are corrected which reduces the dependency upon the batch size.

\section{Conclusion}
In this paper, we propose a gradient norm correction for the adaptive SGD optimizers based on the history of gradient norm. The proposed approach improves the representation of the gradient by boosting its norm to at least the historical gradient norm. The proposed approach is beneficial under flat and high gradient regions to improve the weight updates. The proposed approach is generic. We use it with Adam, diffGrad, Radam and AdaBelief optimizers and propose AdamNorm, diffGradNorm, RadamNorm and AdaBeliefNorm optimizers, respectively. The performance of the optimizers are significantly improved when used with the proposed gradient norm correction approach on CIFAR10, CIFAR100 and TinyImageNet datasets using VGG16, ResNet18 and ResNet50 models. The smaller value of hyperparameter $\gamma$ is better suitable on TinyImageNet dataset. The proposed approach is suitable with only first moment. The smaller learning rate is preferred with the proposed AdamNorm on TinyImageNet dataset. The effect of batch size becomes negligible due to the proposed gradient norm correction.

{\small
\bibliographystyle{ieee_fullname}
\bibliography{References}
}

\section*{Supplementary}

\subsection*{A. Convergence Proof}

\begin{lemma}
\label{lemma:1}
Let $\eta \overset{\triangle}{=} \frac{\beta_1^2}{\sqrt{\beta_2}}$. For $\beta_1$, $\beta_2$ $\in [0,1)$ that satisfy $\frac{\beta_1^2}{\sqrt{\beta_2}} < 1$ and bounded $g_t$, $||g_t||_2 \leq G$, $||g_t||_\infty \leq G_\infty$, $e_t \leq G_\infty$, $\frac{e_t}{||g_t||_2} \leq \frac{G_\infty}{G}$, the following inequality holds,
\begin{dmath*}
\sum_{t=1}^{T}{\frac{\hat{m}_{t,i}^2}{\sqrt{t\hat{v}_{t,i}}}}
\leq
\frac{2G^3_\infty}{G^2(1-\eta)^2\sqrt{1-\beta_2}} ||g_{1:T,i}||_2
\end{dmath*}

\begin{proof}
Under the assumption, $\frac{\sqrt{1-\beta_2^t}}{(1-\beta_1^t)^2} \leq \frac{1}{(1-\beta_1)^2}$. We can use the update rules of AdamNorm and expand the last term in the summation,
\begin{dmath*}
\sum_{t=1}^{T}{\frac{\hat{m}_{t,i}^2}{\sqrt{t\hat{v}_{t,i}}}} = \sum_{t=1}^{T-1}{\frac{\hat{m}_{t,i}^2}{\sqrt{t\hat{v}_{t,i}}}} + \frac{\sqrt{1-\beta_2^T}}{(1-\beta_1^T)^2} \frac{(\sum_{k=1}^{T}{(1-\beta_1)\beta_1^{T-k}s_{k,i}})^2}{\sqrt{T\sum_{j=1}^{T}{(1-\beta_2)\beta_2^{T-j}g_{j,i}^2}}}
\leq 
\sum_{t=1}^{T-1}{\frac{\hat{m}_{t,i}^2}{\sqrt{t\hat{v}_{t,i}}}} + \frac{\sqrt{1-\beta_2^T}}{(1-\beta_1^T)^2} \sum_{k=1}^{T}\frac{T({(1-\beta_1)\beta_1^{T-k}s_{k,i}})^2}{\sqrt{T\sum_{j=1}^{T}{(1-\beta_2)\beta_2^{T-j}g_{j,i}^2}}}
\leq 
\sum_{t=1}^{T-1}{\frac{\hat{m}_{t,i}^2}{\sqrt{t\hat{v}_{t,i}}}} + \frac{\sqrt{1-\beta_2^T}}{(1-\beta_1^T)^2} \sum_{k=1}^{T}\frac{T({(1-\beta_1)\beta_1^{T-k}s_{k,i}})^2}{\sqrt{T(1-\beta_2)\beta_2^{T-k}g_{k,i}^2}}
\end{dmath*}
Further, we can simplify as,
\begin{dmath*}
\sum_{t=1}^{T}{\frac{\hat{m}_{t,i}^2}{\sqrt{t\hat{v}_{t,i}}}} 
\leq 
\sum_{t=1}^{T-1}{\frac{\hat{m}_{t,i}^2}{\sqrt{t\hat{v}_{t,i}}}} + \frac{1}{(1-\beta_1)^2} \sum_{k=1}^{T}\frac{T({(1-\beta_1)\beta_1^{T-k}s_{k,i}})^2}{\sqrt{T(1-\beta_2)\beta_2^{T-k}g_{k,i}^2}}
= 
\sum_{t=1}^{T-1}{\frac{\hat{m}_{t,i}^2}{\sqrt{t\hat{v}_{t,i}}}} + \frac{T}{\sqrt{T(1-\beta_2)}} \sum_{k=1}^{T}\frac{({\beta_1^{T-k}s_{k,i}})^2}{\sqrt{\beta_2^{T-k}g_{k,i}^2}}
= 
\sum_{t=1}^{T-1}{\frac{\hat{m}_{t,i}^2}{\sqrt{t\hat{v}_{t,i}}}} + \frac{T}{\sqrt{T(1-\beta_2)}} \sum_{k=1}^{T}\left(\frac{\beta_1^2}{\sqrt{\beta_2}} \right)^{T-k}\frac{s_{k,i}^2}{g_{k,i}}
= 
\sum_{t=1}^{T-1}{\frac{\hat{m}_{t,i}^2}{\sqrt{t\hat{v}_{t,i}}}} + \frac{T}{\sqrt{T(1-\beta_2)}} \sum_{k=1}^{T}\eta^{T-k}\left(\frac{s_{k,i}}{\sqrt{g_{k,i}}}\right)^2
\leq 
\sum_{t=1}^{T-1}{\frac{\hat{m}_{t,i}^2}{\sqrt{t\hat{v}_{t,i}}}} + \frac{T}{\sqrt{T(1-\beta_2)}} \sum_{k=1}^{T}\eta^{T-k}\left(\frac{\max(1,\frac{e_{k}}{||g_{k}||_2})g_{k,i}}{\sqrt{g_{k,i}}}\right)^2
\end{dmath*}
By considering the bound of $e_k$ and $||g_k||_2$, we can rewrite the above relation as,
\begin{dmath*}
\sum_{t=1}^{T}{\frac{\hat{m}_{t,i}^2}{\sqrt{t\hat{v}_{t,i}}}} 
\leq
\sum_{t=1}^{T-1}{\frac{\hat{m}_{t,i}^2}{\sqrt{t\hat{v}_{t,i}}}} + \frac{T}{\sqrt{T(1-\beta_2)}} \sum_{k=1}^{T}\eta^{T-k}\frac{G^2_\infty}{G^2} ||g_{k,i}||_2
\end{dmath*}
Similarly, after considering the upper bound of the rest of the terms in the summation, we can get as follows,
\begin{dmath*}
\sum_{t=1}^{T}{\frac{\hat{m}_{t,i}^2}{\sqrt{t\hat{v}_{t,i}}}}
\leq
\frac{G^2_\infty}{G^2 \sqrt{(1-\beta_2)}} \sum_{t=1}^{T}\frac{||g_{t,i}||_2}{\sqrt{t}}\sum_{j=0}^{T-t}t\eta^j
\leq
\frac{G^2_\infty}{G^2 \sqrt{(1-\beta_2)}} \sum_{t=1}^{T}\frac{||g_{t,i}||_2}{\sqrt{t}}\sum_{j=0}^{T}t\eta^j
\end{dmath*}
We can obtain $\sum_{t}t\eta^t < \frac{1}{(1-\eta)^2}$ for $\eta < 1$ using the upper bound on the arithmetic-geometric series. Hence,
\begin{dmath*}
\sum_{t=1}^{T}{\frac{\hat{m}_{t,i}^2}{\sqrt{t\hat{v}_{t,i}}}}
\leq
\frac{G^2_\infty}{G^2(1-\eta)^2\sqrt{1-\beta_2}} \sum_{t=1}^{T}\frac{||g_{k,i}||_2}{\sqrt{t}}
\end{dmath*}
By applying Lemma 10.3 of \cite{adam}, we can get,
\begin{dmath*}
\sum_{t=1}^{T}{\frac{\hat{m}_{t,i}^2}{\sqrt{t\hat{v}_{t,i}}}}
\leq
\frac{2G^3_\infty}{G^2(1-\eta)^2\sqrt{1-\beta_2}} ||g_{1:T,i}||_2
\end{dmath*}

\end{proof}
\end{lemma}

\begin{theorem}
\textit{Let the bounded gradients for function $f_t$ (i.e., $||g_{t,\theta}||_2 \leq G$ and $||g_{t,\theta}||_{\infty} \leq G_{\infty}$) for all $\theta \in R^d$. Also assume that AdamNorm produces the bounded distance between any $\theta_t$ (i.e., $||\theta_n-\theta_m||_2 \leq D$ and $||\theta_n-\theta_m||_\infty \leq D_\infty$ for any $m,n\in\{1,...,T\}$). Let $\eta \triangleq \frac{\beta_1^2}{\sqrt{\beta_2}}$, $\beta_1,\beta_2 \in [0,1)$ satisfy $\frac{\beta_1^2}{\sqrt{\beta_2}} < 1$, $\alpha_t=\frac{\alpha}{\sqrt{t}}$, and $\beta_{1,t}=\beta_1\lambda^{t-1},\lambda \in (0,1)$ with $\lambda$ is typically close to $1$, e.g., $1-10^{-8}$. For all $T \geq 1$, the proposed AdamNorm optimizer shows the following guarantee:} 
\begin{dmath*}
R(T) \leq \frac{D^2}{2\alpha(1-\beta_1)}\sum_{i=1}^{d}{\sqrt{T\hat{v}_{T,i}}} 
+ \frac{\alpha(1+\beta_1) G^3_\infty}{(1-\beta_1)\sqrt{1-\beta_2}(1-\eta)^2G^2}\sum_{i=1}^{d}{||g_{1:T,i}||_2} 
+ \sum_{i=1}^{d}{\frac{D_{\infty}^{2}G_{\infty}\sqrt{1-\beta_2}}{2\alpha (1-\beta_1)(1-\lambda)^2}}
\end{dmath*}

\begin{proof}[Proof]
Using Lemma 10.2 of Adam \cite{adam}, we can write as
$$
f_t(\theta_t)-f_t(\theta^*) \leq g_t^T(\theta_t-\theta^*) = \sum_{i=1}^{d}{g_{t,i}(\theta_{t,i}-\theta_{,i}^*)}
$$
We can write following from the AdamNorm update rule, ignoring $\epsilon$, 

\begin{dmath*}
\theta_{t+1} =\theta_t-\frac{\alpha_t\hat{m}_t}{\sqrt[]{\hat{v}_{t}}}
 =\theta_t-\frac{\alpha_t}{(1-\beta_1^t)} \Big(\frac{\beta_{1,t}}{\sqrt[]{\hat{v}_{t}}}m_{t-1} + \frac{(1-\beta_{1,t})}{\sqrt[]{\hat{v}_{t}}}g_t\Big)
\end{dmath*}
where $\beta_{1,t}$ is the $1^{st}$ order moment coefficient at $t^{th}$ iteration and $\beta_1^t$ is the $t^{th}$ power of initial $1^{st}$ order moment coefficient. \\
For $i^{th}$ dimension of parameter vector $\theta_t \in R^d$, we can write
\begin{dmath*}
(\theta_{t+1,i}-\theta_{,i}^*)^2=(\theta_{t,i}-\theta_{,i}^*)^2-\frac{2\alpha_t}{1-\beta_1^t}
\Big(\frac{\beta_{1,t}}{\sqrt[]{\hat{v}_{t,i}}}m_{t-1,i} + \frac{(1-\beta_{1,t})}{\sqrt[]{\hat{v}_{t,i}}}g_{t,i}\Big)(\theta_{t,i}-\theta_{,i}^*)
+\alpha_t^2 (\frac{\hat{m}_{t,i}}{\hat{v}_{t,i}})^2
\end{dmath*}
The above equation can be reordered as
\begin{dmath*}
g_{t,i}(\theta_{t,i}-\theta_{,i}^*)=\frac{(1-\beta_1^t)\sqrt{\hat{v}_{t,i}}}{2\alpha_t(1-\beta_{1,t})}
\Big((\theta_{t,i}-\theta_{,i}^*)^2-(\theta_{t+1,i}-\theta_{,i}^*)^2\Big)
+\frac{\beta_{1,t}}{1-\beta_{1,t}}(\theta_{,i}^*-\theta_{t,i})m_{t-1,i}
+\frac{\alpha_t(1-\beta_1^t)}{2(1-\beta_{1,t})}\frac{(\hat{m}_{t,i})^2}{\sqrt{\hat{v}_{t,i}}}.
\end{dmath*}
Further, it can be written as
\begin{dmath*}
g_{t,i}(\theta_{t,i}-\theta_{,i}^*) =\frac{(1-\beta_1^t)\sqrt{\hat{v}_{t,i}}}{2\alpha_t(1-\beta_{1,t})}
\Big((\theta_{t,i}-\theta_{,i}^*)^2-(\theta_{t+1,i}-\theta_{,i}^*)^2\Big)
 +\sqrt{\frac{\beta_{1,t}}{\alpha_{t-1}(1-\beta_{1,t})}(\theta_{,i}^*-\theta_{t,i})^2\sqrt{\hat{v}_{t-1,i}}} \sqrt{\frac{\beta_{1,t}\alpha_{t-1}(m_{t-1,i})^2}{(1-\beta_{1,t})\sqrt{\hat{v}_{t-1,i}}}}
+\frac{\alpha_t(1-\beta_1^t)}{2(1-\beta_{1,t})}\frac{(\hat{m}_{t,i})^2}{\sqrt{\hat{v}_{t,i}}}
\end{dmath*}
Based on Young's inequality, $ab \leq a^2/2+b^2/2$ and fact that $\beta_{1,t} \leq \beta_1$, the above equation can be reordered as 
\begin{dmath*}
g_{t,i}(\theta_{t,i}-\theta_{,i}^*) \leq \frac{1}{2\alpha_t(1-\beta_1)}
\Big((\theta_{t,i}-\theta_{,i}^*)^2-(\theta_{t+1,i}-\theta_{,i}^*)^2\Big)\sqrt{\hat{v}_{t,i}}
 +\frac{\beta_{1,t}}{2\alpha_{t-1}(1-\beta_{1,t})}(\theta_{,i}^*-\theta_{t,i})^2\sqrt{\hat{v}_{t-1,i}} + \frac{\beta_1\alpha_{t-1}(m_{t-1,i})^2}{2(1-\beta_1)\sqrt{\hat{v}_{t-1,i}}}
+\frac{\alpha_t}{2(1-\beta_1)}\frac{(\hat{m}_{t,i})^2}{\sqrt{\hat{v}_{t,i}}}
\end{dmath*}
We use the Lemma \ref{lemma:1} and derive the regret bound by aggregating it across all the dimensions for $i\in \{1,\dots,d\}$ and all the sequence of convex functions for $t\in \{1,\dots,T\}$ in the upper bound of $f_t(\theta_t)-f_t(\theta^*)$ as
\begin{dmath*}
R(T) \leq \sum_{i=1}^{d}{\frac{1}{2\alpha_1(1-\beta_1)}} (\theta_{1,i}-\theta_{,i}^*)^2\sqrt{\hat{v}_{1,i}} + \sum_{i=1}^{d}{\sum_{t=2}^{T}{\frac{1}{2(1-\beta_1)}} (\theta_{t,i}-\theta_{,i}^*)^2(\frac{\sqrt{\hat{v}_{t,i}}}{\alpha_t}-\frac{\sqrt{\hat{v}_{t-1,i}}}{\alpha_{t-1}})}
 + \frac{\beta_1\alpha G^3_\infty}{(1-\beta_1)\sqrt{1-\beta_2}(1-\eta)^2G^2}\sum_{i=1}^{d}{||g_{1:T,i}||_2}
+ \frac{\alpha G^3_\infty}{(1-\beta_1)\sqrt{1-\beta_2}(1-\eta)^2G^2}\sum_{i=1}^{d}{||g_{1:T,i}||_2}
 + \sum_{i=1}^{d}{\sum_{t=1}^{T}{\frac{\beta_{1,t}}{2\alpha_{t}(1-\beta_{1,t})}(\theta_{,i}^*-\theta_{t,i})^2\sqrt{\hat{v}_{t,i}}}}
\end{dmath*}
By utilizing the assumptions that $\alpha=\alpha_t\sqrt{t}$, $||\theta_t-\theta^*||_2 \leq D$ and $||\theta_m-\theta_n||_{\infty} \leq D_{\infty}$, we can write as
\begin{dmath*}
R(T) \leq \frac{D^2}{2\alpha(1-\beta_1)}\sum_{i=1}^{d}{\sqrt{T\hat{v}_{T,i}}} 
+ \frac{\alpha(1+\beta_1) G^3_\infty}{(1-\beta_1)\sqrt{1-\beta_2}(1-\eta)^2G^2}\sum_{i=1}^{d}{||g_{1:T,i}||_2}
+ \frac{D_{\infty}^{2}}{2\alpha}\sum_{i=1}^{d}{\sum_{t=1}^{t}{\frac{\beta_{1,t}}{(1-\beta_{1,t})}\sqrt{t\hat{v}_{t,i}}}}
  \leq \frac{D^2}{2\alpha(1-\beta_1)}\sum_{i=1}^{d}{\sqrt{T\hat{v}_{T,i}}} 
+ \frac{\alpha(1+\beta_1) G^3_\infty}{(1-\beta_1)\sqrt{1-\beta_2}(1-\eta)^2G^2}\sum_{i=1}^{d}{||g_{1:T,i}||_2} 
+ \frac{D_{\infty}^{2}G_{\infty}\sqrt{1-\beta_2}}{2\alpha}\sum_{i=1}^{d}{\sum_{t=1}^{t}{\frac{\beta_{1,t}}{(1-\beta_{1,t})}\sqrt{t}}}
\end{dmath*}
It is shown in Adam \cite{adam} that $\sum_{t=1}^{t}{\frac{\beta_{1,t}}{(1-\beta_{1,t})}\sqrt{t}} \leq \frac{1}{(1-\beta_1)(1-\eta)^2}$. Thus, the regret bound can be written as
\begin{dmath*}
R(T) \leq \frac{D^2}{2\alpha(1-\beta_1)}\sum_{i=1}^{d}{\sqrt{T\hat{v}_{T,i}}} 
+ \frac{\alpha(1+\beta_1) G^3_\infty}{(1-\beta_1)\sqrt{1-\beta_2}(1-\eta)^2G^2}\sum_{i=1}^{d}{||g_{1:T,i}||_2} 
+ \sum_{i=1}^{d}{\frac{D_{\infty}^{2}G_{\infty}\sqrt{1-\beta_2}}{2\alpha (1-\beta_1)(1-\lambda)^2}}
\end{dmath*}
\end{proof}
\end{theorem}

\subsection*{B. Algorithms}
This section provides the Algorithms for different optimization techniques, including diffGrad (Algorithm \ref{algo:diffgrad}), diffGradInject (Algorithm \ref{algo:diffgradnorm}), Radam (Algorithm \ref{algo:radam}), RadamInject (Algorithm \ref{algo:radamnorm}), AdaBelief (Algorithm \ref{algo:adabelief}) and AdaBeliefInject (Algorithm \ref{algo:adabeliefnorm}).

\begin{algorithm}[!h]
\caption{diffGrad Optimizer}
\SetAlgoLined
\textbf{Initialize:} $\bm{\theta}_{0},\bm{m}_{0}\gets0,\bm{v}_{0}\gets0,t\gets0$\\
\textbf{Hyperparameters:} $\alpha, \beta_1, \beta_2$\\
\textbf{While} $\bm{\theta}_{t}$ not converged\\
    \hspace{0.45cm} $t \gets t+1$\\
    \hspace{0.45cm} $\bm{g}_t \gets \nabla_{\theta} f_t(\bm{\theta}_{t-1})$ \\
    \hspace{0.45cm} $\bm{\xi}_t \gets 1/(1+e^{-|\bm{g}_t - \bm{g}_{t-1}|})$ \\
    \hspace{0.45cm} $\bm{m}_t \gets \beta_1 \bm{m}_{t-1} + (1-\beta_1) \bm{g}_t$\\
    \hspace{0.45cm} $\bm{v}_t \gets \beta_2 \bm{v}_{t-1} + (1-\beta_2) \bm{g}^2_t$\\
    \hspace{0.45cm} \textbf{Bias Correction}\\
    \hspace{0.9cm} $\widehat{\bm{m}_t} \gets \bm{m}_t/(1-\beta_1^t)$, $\widehat{\bm{v}_t} \gets \bm{v}_t/(1-\beta_2^t)$\\
    \hspace{0.45cm} \textbf{Update}\\
    \hspace{0.9cm} $\bm{\theta}_t \gets \bm{\theta}_{t-1} - \alpha \bm{\xi}_t \widehat{\bm{m}_t}/(\sqrt{\widehat{\bm{v}_t}} + \epsilon)$
\label{algo:diffgrad}
\end{algorithm}

\begin{algorithm}[!t]
\caption{diffGradNorm (diffGrad + AdaNorm) Optimizer}
\SetAlgoLined
\textbf{Initialize:} $\bm{\theta}_{0},\bm{m}_{0}\gets0,\bm{v}_{0}\gets0,e_{0}\gets0,t\gets0$\\
\textbf{Hyperparameters:} $\alpha, \beta_1, \beta_2, \gamma$\\
\textbf{While} $\bm{\theta}_{t}$ not converged\\
    \hspace{0.45cm} $t \gets t+1$\\
    \hspace{0.45cm} $\bm{g}_t \gets \nabla_{\theta} f_t(\bm{\theta}_{t-1})$ \\
    \hspace{0.45cm} $\bm{\xi}_t \gets 1/(1+e^{-|\bm{g}_t - \bm{g}_{t-1}|})$ \\
    \hspace{0.45cm} \textcolor{blue}{$g_{norm} \gets L_2Norm(\bm{g}_t)$}\\
    \hspace{0.45cm} \textcolor{blue}{$e_t = \gamma e_{t-1} + (1-\gamma) g_{norm}$}\\
    \hspace{0.45cm} \textcolor{blue}{$\bm{s}_t = \bm{g}_t$}\\
    \hspace{0.45cm} \textcolor{blue}{\textbf{If} $e_t > g_{norm}$}\\
    \hspace{0.9cm} \textcolor{blue}{$\bm{s}_t = (e_t / g_{norm})\bm{g}_t$}\\
    % \hspace{0.45cm} \textbf{Else}\\
    % \hspace{0.90cm} $\bm{s}_t = \bm{g}_t$\\
    \hspace{0.45cm} $\bm{m}_t \gets \beta_1 \bm{m}_{t-1} + (1-\beta_1) \textcolor{blue}{\bm{s}_t}$\\
    \hspace{0.45cm} $\bm{v}_t \gets \beta_2 \bm{v}_{t-1} + (1-\beta_2) \bm{g}^2_t$\\
    \hspace{0.45cm} \textbf{Bias Correction}\\
    \hspace{0.9cm} $\widehat{\bm{m}_t} \gets \bm{m}_t/(1-\beta_1^t)$, $\widehat{\bm{v}_t} \gets \bm{v}_t/(1-\beta_2^t)$\\
    \hspace{0.45cm} \textbf{Update}\\
    \hspace{0.9cm} $\bm{\theta}_t \gets \bm{\theta}_{t-1} - \alpha \bm{\xi}_t \widehat{\bm{m}_t}/(\sqrt{\widehat{\bm{v}_t}} + \epsilon)$
\label{algo:diffgradnorm}
\end{algorithm}

\begin{algorithm}[!t]
\caption{Radam Optimizer}
\SetAlgoLined
\textbf{Initialize:} $\bm{\theta}_{0},\bm{m}_{0}\gets0,\bm{v}_{0}\gets0,t\gets0$\\
\textbf{Hyperparameters:} $\alpha, \beta_1, \beta_2$\\
\textbf{While} $\bm{\theta}_{t}$ not converged\\
    \hspace{0.45cm} $t \gets t+1$\\
    \hspace{0.45cm} $\bm{g}_t \gets \nabla_{\theta} f_t(\bm{\theta}_{t-1})$ \\
    \hspace{0.45cm} $\bm{m}_t \gets \beta_1   \bm{m}_{t-1} + (1-\beta_1)   \bm{g}_t$\\
    \hspace{0.45cm} $\bm{v}_t \gets \beta_2   \bm{v}_{t-1} + (1-\beta_2)   \bm{g}_t^2$\\
    \hspace{0.45cm} $\rho_\infty \gets 2 / (1 - \beta_2) - 1$ \\
    \hspace{0.45cm} $\rho_t = \rho_\infty - 2t\beta_2^t/(1-\beta_2^t)$\\
    \hspace{0.45cm} \textbf{If} $\rho_t \geq 5$\\
    \hspace{0.9cm} $\rho_u = (\rho_t - 4)    (\rho_t - 2)    \rho_\infty$\\
    \hspace{0.9cm} $\rho_d = (\rho_\infty - 4)    (\rho_\infty - 2)    \rho_t$\\
    \hspace{0.9cm} $\rho = \sqrt{(1 - \beta_2)    \rho_u / \rho_d}$\\
    \hspace{0.9cm} $\alpha_1 = \rho    \alpha / (1 - \beta_1^t)$\\
    \hspace{0.9cm} \textbf{Update}\\
    \hspace{1.35cm} $\bm{\theta}_t \gets \bm{\theta}_{t-1} - \alpha_1    \bm{m}_t/(\sqrt{\bm{v}_t} + \epsilon)$ \\
    \hspace{0.45cm} \textbf{Else}\\
    \hspace{0.9cm} $\alpha_2 = \alpha / (1 - \beta_1^t)$\\
    \hspace{0.9cm} \textbf{Update}\\
    \hspace{1.35cm} $\bm{\theta}_t \gets \bm{\theta}_{t-1} - \alpha_2    \bm{m}_t$
\label{algo:radam}
\end{algorithm}

\begin{algorithm}[!t]
\caption{RadamNorm (i.e., Radam + AdaNorm) Optimizer}
\SetAlgoLined
\textbf{Initialize:} $\bm{\theta}_{0},\bm{m}_{0}\gets0,\bm{v}_{0}\gets0,e_0\gets0,t\gets0$\\
\textbf{Hyperparameters:} $\alpha, \beta_1, \beta_2, \gamma$\\
\textbf{While} $\bm{\theta}_{t}$ not converged\\
    \hspace{0.45cm} $t \gets t+1$\\
    \hspace{0.45cm} $\bm{g}_t \gets \nabla_{\theta} f_t(\bm{\theta}_{t-1})$ \\
    \hspace{0.45cm} \textcolor{blue}{$g_{norm} \gets L_2Norm(\bm{g}_t)$}\\
    \hspace{0.45cm} \textcolor{blue}{$e_t = \gamma e_{t-1} + (1-\gamma) g_{norm}$}\\
    \hspace{0.45cm} \textcolor{blue}{$\bm{s}_t = \bm{g}_t$}\\
    \hspace{0.45cm} \textcolor{blue}{\textbf{If} $e_t > g_{norm}$}\\
    \hspace{0.9cm} \textcolor{blue}{$\bm{s}_t = (e_t / g_{norm})\bm{g}_t$}\\
    % \hspace{0.45cm} \textbf{Else}\\
    % \hspace{0.90cm} $\bm{s}_t = \bm{g}_t$\\
    \hspace{0.45cm} $\bm{m}_t \gets \beta_1   \bm{m}_{t-1} + (1-\beta_1)   \textcolor{blue}{\bm{s}_t}$\\
    \hspace{0.45cm} $\bm{v}_t \gets \beta_2   \bm{v}_{t-1} + (1-\beta_2)   \bm{g}_t^2$\\
    \hspace{0.45cm} $\rho_\infty \gets 2 / (1 - \beta_2) - 1$ \\
    \hspace{0.45cm} $\rho_t = \rho_\infty - 2t\beta_2^t/(1-\beta_2^t)$\\
    \hspace{0.45cm} \textbf{If} $\rho_t \geq 5$\\
    \hspace{0.9cm} $\rho_u = (\rho_t - 4)    (\rho_t - 2)    \rho_\infty$\\
    \hspace{0.9cm} $\rho_d = (\rho_\infty - 4)    (\rho_\infty - 2)    \rho_t$\\
    \hspace{0.9cm} $\rho = \sqrt{(1 - \beta_2)    \rho_u / \rho_d}$\\
    \hspace{0.9cm} $\alpha_1 = \rho    \alpha / (1 - \beta_1^t)$\\
    \hspace{0.9cm} \textbf{Update}\\
    \hspace{1.35cm} $\bm{\theta}_t \gets \bm{\theta}_{t-1} - \alpha_1    \bm{m}_t/(\sqrt{\bm{v}_t} + \epsilon)$ \\
    \hspace{0.45cm} \textbf{Else}\\
    \hspace{0.9cm} $\alpha_2 = \alpha / (1 - \beta_1^t)$\\
    \hspace{0.9cm} \textbf{Update}\\
    \hspace{1.35cm} $\bm{\theta}_t \gets \bm{\theta}_{t-1} - \alpha_2    \bm{m}_t$
\label{algo:radamnorm}
\end{algorithm}

\begin{algorithm}[!t]
\caption{AdaBelief Optimizer}
\SetAlgoLined
\textbf{Initialize:} $\bm{\theta}_{0},\bm{m}_{0}\gets0,\bm{v}_{0}\gets0,t\gets0$\\
\textbf{Hyperparameters:} $\alpha, \beta_1, \beta_2$\\
\textbf{While} $\bm{\theta}_{t}$ not converged\\
    \hspace{0.45cm} $t \gets t+1$\\
    \hspace{0.45cm} $\bm{g}_t \gets \nabla_{\theta} f_t(\bm{\theta}_{t-1})$ \\
    \hspace{0.45cm} $\bm{m}_t \gets \beta_1   \bm{m}_{t-1} + (1-\beta_1)   \bm{g}_t$\\
    \hspace{0.45cm} $\bm{v}_t \gets \beta_2   \bm{v}_{t-1} + (1-\beta_2)   (\bm{g}_t - \bm{m}_t)^2$\\
    \hspace{0.45cm} \textbf{Bias Correction}\\
    \hspace{0.9cm} $\widehat{\bm{m}_t} \gets \bm{m}_t/(1-\beta_1^t)$, $\widehat{\bm{v}_t} \gets \bm{v}_t/(1-\beta_2^t)$\\
    \hspace{0.45cm} \textbf{Update}\\
    \hspace{0.9cm} $\bm{\theta}_t \gets \bm{\theta}_{t-1} - \alpha \widehat{\bm{m}_t}/(\sqrt{\widehat{\bm{v}_t}} + \epsilon)$
\label{algo:adabelief}    
\end{algorithm}

\begin{algorithm}[!t]
\caption{AdaBeliefNorm (AdaBelief + AdaNorm) Optimizer}
\SetAlgoLined
\textbf{Initialize:} $\bm{\theta}_{0},\bm{m}_{0}\gets0,\bm{v}_{0}\gets0,e_0\gets0,t\gets0$\\
\textbf{Hyperparameters:} $\alpha, \beta_1, \beta_2,\gamma$\\
\textbf{While} $\bm{\theta}_{t}$ not converged\\
    \hspace{0.45cm} $t \gets t+1$\\
    \hspace{0.45cm} $\bm{g}_t \gets \nabla_{\theta} f_t(\bm{\theta}_{t-1})$ \\
    \hspace{0.45cm} \textcolor{blue}{$g_{norm} \gets L_2Norm(\bm{g}_t)$}\\
    \hspace{0.45cm} \textcolor{blue}{$e_t = \gamma e_{t-1} + (1-\gamma) g_{norm}$}\\
    \hspace{0.45cm} \textcolor{blue}{$\bm{s}_t = \bm{g}_t$}\\
    \hspace{0.45cm} \textcolor{blue}{\textbf{If} $e_t > g_{norm}$}\\
    \hspace{0.9cm} \textcolor{blue}{$\bm{s}_t = (e_t / g_{norm})\bm{g}_t$}\\
    % \hspace{0.45cm} \textbf{Else}\\
    % \hspace{0.90cm} $\bm{s}_t = \bm{g}_t$\\
    \hspace{0.45cm} $\bm{m}_t \gets \beta_1   \bm{m}_{t-1} + (1-\beta_1)   \textcolor{blue}{\bm{s}_t}$\\
    \hspace{0.45cm} $\bm{v}_t \gets \beta_2   \bm{v}_{t-1} + (1-\beta_2)   (\bm{g}_t - \bm{m}_t)^2$\\
    \hspace{0.45cm} \textbf{Bias Correction}\\
    \hspace{0.9cm} $\widehat{\bm{m}_t} \gets \bm{m}_t/(1-\beta_1^t)$, $\widehat{\bm{v}_t} \gets \bm{v}_t/(1-\beta_2^t)$\\
    \hspace{0.45cm} \textbf{Update}\\
    \hspace{0.9cm} $\bm{\theta}_t \gets \bm{\theta}_{t-1} - \alpha \widehat{\bm{m}_t}/(\sqrt{\widehat{\bm{v}_t}} + \epsilon)$
\label{algo:adabeliefnorm}
\end{algorithm}

\end{document}